\documentclass[journal]{IEEEtai}

\usepackage[colorlinks,urlcolor=blue,linkcolor=blue,citecolor=blue]{hyperref}

\usepackage{color,array}

\usepackage{graphicx}
\usepackage{amssymb}
\usepackage{amsmath}
\usepackage{tabularx,caption}
\usepackage{booktabs,multirow,array}
\usepackage{siunitx} 

\begin{document}

\title{Dual Attention U-Net with Feature Infusion: Pushing the Boundaries of Multiclass Defect Segmentation}

\author{Rasha Alshawi, Md Tamjidul Hoque, Md Meftahul Ferdaus, Mahdi Abdelguerfi, Kendall Niles, Ken Prathak, Joe Tom, Jordan Klein, Murtada Mousa, and Johny Javier Lopez 
\thanks{``This research was supported in part by the U.S. Department of the Army – U.S. Army Corps of Engineers (USACE) under contract W912HZ-23-2-0004. The views expressed in this paper are solely those of the authors and do not necessarily reflect the views of the funding agencies.''}
\thanks{Rasha Alshawi, Md Tamjidul Hoque, Md Meftahul Ferdaus, Mahdi Abdelguerfi, and Johny Javier Lopez are with the Canizaro Livingston Gulf States Center for Environmental Informatics, Department of Computer Science, The University of New Orleans, LA 70148 USA (e-mail: ralshawi@uno.edu, thoque@uno.edu, mferdaus@uno.edu., mahdi@cs.uno.edu)}
\thanks{Kendall Niles, Ken Prathak, Joe Tom, and Jordan Klein are with US Army Corps of Engineers, Vicksburg, Mississippi 39183 USA (e-mail: Kendall.N.Niles@erdc.dren.mil).}
\thanks{Murtada Mousa is registered professional engineer in the State of Louisiana (e-mail: murtadamousa@providenceeng.com).}
\thanks{This paragraph will include the Associate Editor who handled your paper.}}

\markboth{Journal of IEEE Transactions on Artificial Intelligence, Vol. 00, No. 0, Month 2024}
{First A. Author \MakeLowercase{\textit{et al.}}: Bare Demo of IEEEtai.cls for IEEE Journals of IEEE Transactions on Artificial Intelligence}

\maketitle

\begin{abstract}
The proposed architecture, Dual Attentive U-Net with Feature Infusion (DAU-FI Net), addresses challenges in semantic segmentation, particularly on multiclass imbalanced datasets with limited samples. DAU-FI Net integrates multiscale  spatial-channel attention mechanisms and feature injection to enhance precision in object localization. The core employs a multiscale depth-separable convolution block, capturing localized patterns across scales. This block is complemented by a spatial-channel squeeze and excitation (scSE) attention unit, modeling inter-dependencies between channels and spatial regions in feature maps. Additionally, additive attention gates refine segmentation by connecting encoder-decoder pathways. 

To augment the model, engineered features using Gabor filters for textural analysis, Sobel and Canny filters for edge detection are injected guided by semantic masks to expand the feature space strategically. Comprehensive experiments on a challenging sewer pipe and culvert defect dataset and a benchmark dataset validate DAU-FI Net’s capabilities. Ablation studies highlight incremental benefits from attention blocks and feature injection. DAU-FI Net achieves state-of-the-art mean Intersection over Union (IoU) of 95.6\% and 98.8\% on the defect test set and benchmark respectively, surpassing prior methods by 8.9\% and 12.6\%, respectively. Ablation studies highlight incremental benefits from attention blocks and feature injection. The proposed architecture provides a robust solution, advancing semantic segmentation for multiclass problems with limited training data. Our sewer-culvert defects dataset, featuring pixel-level annotations, opens avenues for further research in this crucial domain. Overall, this work delivers key innovations in architecture, attention, and feature engineering to elevate semantic segmentation efficacy.
\end{abstract}

\begin{IEEEImpStatement}
Automatic defect detection in underground infrastructure is crucial yet challenging, with manual inspection being hazardous, time-consuming, and prone to errors. Our proposed DAU-FI Net architecture provides an accurate automated solution, overcoming key limitations of prior methods. By integrating innovations in attention mechanisms and feature engineering, DAU-FI Net achieved 95.6\% and 98.8\% intersection-over-union on a sewer-culvert defect dataset and cell nuclei benchmark, respectively, surpassing state-of-the-art by 8.9\% and 12.6\%. This level of performance on complex real-world data enables reliable automation of infrastructure inspection, improving efficiency and safety. The curated pixel-annotated defect dataset also opens up new research avenues. With further development, this technology can be deployed in the field to autonomously analyze sewers, tunnels, and pipelines, providing rapid anomaly detection to prioritize maintenance and prevent catastrophic failures. This work establishes an integral benchmark, pushing boundaries in architectural design, attention modeling, and feature infusion to elevate deep learning capabilities.
\end{IEEEImpStatement}

\begin{IEEEkeywords}
Attention Mechanism, Feature Engineering, Infrastructure Inspection, Semantic Segmentation, U-Net
\end{IEEEkeywords}

\section{Introduction}\label{sec:intro}

\IEEEPARstart{S}{emantic} segmentation, a vital subfield of computer vision, continues to be an active and dynamic area of research. This technique enables machines to better understand visual scenes by labeling each pixel with a corresponding class and precise spatial location. Among the most promising advances in this domain are fully convolutional networks (FCNs). FCNs excel at semantic segmentation through an elegant encoder-decoder architecture centered around convolutional neural networks (CNNs). The encoder portion condenses an input image into a compact latent representation via convolutional and downsampling layers. This representation encoding high-level features is then passed to the decoder network, which up-samples to produce a segmentation map matching the original image dimensions. The encoder-decoder structure allows FCNs to perform end-to-end pixel-wise prediction for inputs of arbitrary sizes.

FCNs are comprehensive neural network architectures that can accommodate inputs of varying dimensions and generate outputs with matched dimensions. This unique architectural approach enables FCNs to efficiently assign categorical labels to individual pixels within an image, aligning them with their corresponding object categories \cite{roy2018recalibrating}. A particularly noteworthy FCN architecture is U-Net, which has recently gained prominence, especially for its ability to produce precise segmentation outputs even when trained on limited data \cite{ronneberger2015u}. The U-Net's encoder-decoder structure and incorporation of skip connections allow it to leverage both local details and global context when generating segmentations. By combining convolutional layers that capture hierarchical feature representations with upsampling layers that enable precise localization, U-Nets achieve excellent performance in medical imaging and other segmentation tasks requiring detailed delineation of complex structures. The network's capacity to yield accurate output from small datasets has been invaluable for many applications where large-scale annotated training data are difficult to obtain.

Building upon recent advancements in semantic segmentation leveraging FCNs and U-Net architectures, our own prior work aimed to further enhance segmentation performance. In our preceding work \cite{alshawi2023depth}, we introduced an enhanced iteration of the U-Net architecture, which incorporated a depth-wise separable block enriched with multiscale filters. The multiscale depth separable block encompasses a sequence of separable convolutions, featuring diverse kernel dimensions ($3\times3$ and $5\times5$), alongside a $1\times1$ convolution with diminished filters. This combination of convolutions captures a spectrum of local patterns and extracts features spanning multiple scales. While the model demonstrated promising performance, further advancements are needed to address multiclass datasets' limitations and challenges, particularly in class imbalance instances and with constricted sample availability. Additionally, differentiating classes that resemble each other within the dataset remains challenging.

This paper addresses the aforementioned challenges by introducing different attention mechanisms into our model. Specifically, we propose combining the multiscale depth separable block with an enhanced squeeze-and-excitation (SE) mechanism. The primary objective of this combined block is to enhance the model's representational capabilities by capturing both spatial and channel-wise dependencies within the input data. This dual attention approach allows the model to incorporate localized patterns and global context to improve segmentation performance.

Furthermore, we introduce a novel strategy that integrates extracted image features into the model to enhance performance. Although feature engineering became less crucial with the rise of deep learning, combining handcrafted features with deep neural networks remains advantageous, particularly in scenarios involving limited training data or few semantic classes. This hybrid approach can effectively expand the feature space and offer supplementary information to guide the model's learning process. 

To evaluate the performance of our model, we created a dedicated dataset for segmenting deficiencies in sewer pipes and culverts containing images of various defect types identified and annotated by domain experts. The deficiencies include cracks, fractures, pipe deformations, joint issues, and other complex damage modalities commonly encountered during sewer and culvert inspections. Precise pixel-level annotations were obtained by manually tracing the boundaries of deficiencies in each image using the LabelMe annotation tool. Furthermore, to evaluate the generalizability of the model, it was tested against the 2018 Data Science Bowl dataset for cell nuclei segmentation \cite{caicedo2019nucleus}, which is recognized as a public benchmark. This benchmark comprises a diverse array of 2D light microscopy images with segmentation ground truths. The dual nature of this evaluation—focusing both on the specific task of segmenting fine-grained pipe deficiencies and on a wider range of segmentation challenges, allows for a comprehensive analysis of the model’s capabilities.

The creation of a custom dataset for the segmentation of sewer-culvert defects, which includes the challenging characteristics observed in practical situations, not only facilitates an in-depth evaluation of our model but also serves as a valuable asset to advance future research in this critical field.

\subsection{Main Contributions}
This work makes several key contributions advancing semantic segmentation for multiclass problems with limited training data:

\begin{itemize}
    \item We propose a novel dual attentive U-Net architecture (DAU-FI Net) that integrates customized multiscale spatial-channel attention mechanisms and strategically infuses engineered image features to enhance precision for multiclass segmentation with limited training data. It introduces a dual attentive block that fuses multiscale convolutions and concurrent spatial-channel squeeze and excitation modeling to capture both localized patterns and global context.
    \item We achieve state-of-the-art performance on a challenging real-world sewer-culvert defect segmentation dataset collected and annotated here. It outperforms prior methods by significant margins and validates generalization capabilities on a cell nuclei segmentation benchmark. Detailed ablation studies analyze the incremental benefits of key components.
    \item We provide an impactful sewer-culvert defect dataset with pixel-level annotations spanning diverse deficiency types to advance future research in this safety-critical domain. In-depth ablation studies analyze how attention mechanisms and strategic feature infusion incrementally improve multiclass segmentation with scarce training data. The code of our proposed models are publicly accessible through the following link: \href{https://github.com/RashaAlshawi/Dual-Attention-U-Net-with-Feature-Infusion-Pushing-the-Boundaries-of-Multiclass-Defect-Segmentation.git}{https://tinyurl.com/DAUFINet}
    
\end{itemize}

\section{Literature Review on the Evolution of Semantic Segmentation}
In recent years, numerous semantic segmentation models have proposed, each aiming to outperform predecessors through diverse strategies and techniques. A transformative development is the integration of attention mechanisms into segmentation models. By focusing on salient regions while suppressing irrelevant details, attention modules replicate integral aspects of human visual perception. When effectively incorporated, attention enhances the performance of models, contributing to advancements in the field.  This section reviews seminal segmentation models, analyzes key limitations motivating our approach, and introduce relevant attention mechanisms.

At its core, semantic segmentation categorically labels each pixel in an image via pixel-level classification, commonly achieved using \textbf{fully convolutional neural networks (FCNNs)} in an encoder-decoder architecture. The encoder condenses spatial dimensions through successive convolutional and downsampling layers, compressing the input into compact latent representation encoding salient features. The decoder then upsamples this encoding to produce a segmentation map matching the original input resolution. Iterations of this overall design have become well-known, consistently improving on earlier methods. For example, the FCN-8s model \cite{long2015fully} utilizes skip connections to recover fine-grained spatial details lost during encoding, improving segmentation precision on datasets like PASCAL VOC \cite{long2015fully}.

Roy et al. proposed the Spatial and Channel Squeeze \& Excitation (scSE) model, representing a targeted advancement of the conventional squeeze-and-excitation block to enhance fully convolutional networks for semantic segmentation \cite{roy2018recalibrating}. This is achieved by recalibrating channel-wise feature responses to capture inter-dependencies within feature maps via learnable weight layers. ScSE dynamically recalibrates activations during the forward pass, up-weighting informative features while suppressing less useful ones. Empirically, integrating scSE improves segmentation accuracy and object delineation by enhancing representational power and modeling richer channel inter-relationships. However, potential limitations include increased computational overhead from added model complexity and needing comprehensive analysis of adaptability across diverse architectures and tasks.

Complementing these advances, Su et al. investigated integrating lightweight Convolutional Block Attention Modules (CBAM) into U-Nets \cite{su2022research}. The Channel Attention Module (CAM) operates along the channel dimension, aggregating features via average pooling and using learnable layers to selectively emphasize informative channels. Meanwhile, the Spatial Attention Module (SAM) generates 2D attention maps to highlight salient spatial regions and objects. This synchronized combination of channel and spatial attention improves segmentation by capturing global semantics and localizing detailed structures. Evaluations demonstrate CBAM’s efficacy, with the dual attention mechanisms boosting representational power and precision of pixel-level object delineation. The modular nature also allows flexible integration into architectures like U-Net without excessive computational overhead.

Building upon fundamental encoder-decoder architectures, the Attention Sparse Convolutional U-Net (ASCU-Net) proposes an innovative tripartite attention scheme to boost semantic segmentation performance \cite{tong2021ascu}. This is achieved by effectively integrating three complementary attention modules - the Attention Gate (AG), Spatial Attention Module (SAM), and Channel Attention Module (CAM). The AG dynamically focuses on important target structures in each encoder layer. It filters out irrelevant regions before transmitting information to the decoder. The lightweight SAM utilizes normalized convolutions to model spatial relationships and localize salient regions. Meanwhile, the CAM adaptively recalibrates channel-wise features via squeeze-and-excite operations to emphasize informative channels. This comprehensive attention mechanism selectively filters and emphasizes the most crucial information, improving representational quality. Their experiments demonstrate state-of-the-art segmentation accuracy, with the concerted attention approach boosting precision and robustness. Elegant integration of customized attention modules represents a significant step forward in the advancement of deep learning techniques for scene understanding.

While prior innovations have contributed unique strengths and pushed boundaries in semantic segmentation, persistent challenges remain in dealing with complex multiclass datasets suffering from class imbalance. To overcome these obstacles, we propose a novel model through an integrated approach.

Our model incorporates various attention mechanisms, including multiscale filtering, to enhance resolution and refine segmentation. We also strategically inject engineered features to expand representational capacity. This balanced combination of top-down attention and bottom-up feature design enhances the effectiveness of segmentation. Together, these innovations have the potential to transform the segmentation of complex real-world scenes such as the inspection of underground infrastructure. Our approach focuses on advancing semantic segmentation research with the goal of developing more reliable and accurate solutions that are readily deployable. By addressing current limitations, our work aims to propel scene understanding to the next level through a flexible architecture specifically designed to handle the challenges posed by imbalanced data across multiple classes.

Automated inspection of underground infrastructure, such as culverts and sewer pipes, is vital for identifying structural and material degradation. This ensures proper function throughout the design life. Automated defect detection is crucial for improving infrastructure owners' ability to make data-driven maintenance decisions and mitigating personnel risks related to health and safety. However, it comes with challenges as manual inspection is slow, costly, and error-prone \cite{pan2020automatic}. These environments exhibit poor lighting, occlusions, and critically, a diverse array of defect types including cracks, corrosion, blockages, joint issues, and intrusions \cite{alshawi2023depth,koch2015review}. This heterogeneity of defects significantly complicates analysis.

Furthermore, as highlighted by Haurum et al. \cite{haurum2021sewerml} and Gao et al. \cite{gao2021decision}, additional difficulties arise, such as limited lighting, textural variations, and occlusions from water and debris. Class imbalance also occurs, with underrepresented defects. Together, these issues make precise identification and categorization exceptionally difficult.

However, recent literature has seen advancements in detecting cracks in infrastructure using deep learning and machine learning. Panta et al. \cite{panta2023} introduced IterLUNet, an encoder-decoder network for pixel-level crack detection in levee images. A comparative study \cite{panta2022pixel} found MultiResUnet achieved the highest mean Intersection over Union. Continuous monitoring was emphasized, given the catastrophic risks posed by cracks \cite{panta2022}. Prior research has assessed algorithms for crack detection from imagery \cite{kuchi2021,kuchi2020} and pioneered sand boil detection using machine learning \cite{kuchi2019}.

These studies demonstrate progress in deploying sophisticated algorithms for detecting structural weaknesses, aiding disaster prevention. However, challenges remain regarding complex real-world environments and aerial inspection data limitations.

Our model aims to address these challenges through customized attention mechanisms and expansive feature representations, enabling effective segmentation even with imbalanced, limited data. Strategic feature injection further strengthens capabilities by integrating fundamental image processing and learned representations. Our approach facilitates efficient, dependable automation of this traditionally manual process

\section{Model Architecture of Dual Attentive U-Net with Feature Infusion}
This section details the architecture of our proposed model, Dual Attentive U-Net with Feature Infusion (DAU-FI Net), as depicted in Fig. \ref{fig:whole_archi}. The model contains several key components that enable enhanced semantic segmentation capabilities through dual attention modeling and strategic feature engineering:

\begin{itemize}
    \item The Dual Attentive Block (DAB) attention mechanism is shown on the left of Fig. \ref{fig:whole_archi}.
    \item The attention gate, displayed on the top right of Fig. \ref{fig:whole_archi}.
    \item Attentive skip connections between the encoder and decoder pathways, enclosed within the bottom right block of Fig. \ref{fig:whole_archi}.
    \item The feature augmentation pipeline is enclosed within the bottom right block of Fig. \ref{fig:whole_archi}.
\end{itemize}

In the next section, we will start by explaining the second and third bullet points above to understand the U-Net encoder-decoder backbone with attentive skip connections. Details on the DAB architecture and the feature augmentation pipeline will follow this.

\subsection{U-Net Encoder-Decoder Backbone with Attentive Skip Connections}
The DAU-FI Net implements an attentive encoder-decoder architecture for precise semantic segmentation, building upon our previous refined U-Net architecture \cite{alshawi2023depth}. This U-shaped topology integrates innovations including attention gates, skip connections, and coordinated upsampling/downsampling between the encoder and decoder paths, as indicated by the data flow through DABs and concatenation nodes. Max pooling downsamples in the encoder path, while transpose convolutions (up-convolutions) upsample in the decoder path. Together, these components boost modeling capabilities for improved segmentation.

A key innovation in the DAU-FI Net is the integration of additive attention gates (top right of Fig.\ref{fig:whole_archi}) into the skip connections linking the encoder and decoder pathways. These attention gates act as specialized filters to selectively emphasize the most salient structures in the encoder feature maps while suppressing irrelevant regions. As data flows through the attention gates, they adaptively aggregate spatial contexts, allowing only the most useful information to pass through. This attention-guided communication between the encoder and decoder pathways boosts the precision of the model's segmentation capabilities. The attention gates are strategically positioned before concatenation nodes, where encoder and decoder features are combined. This mechanism enables the gates to prioritize feature maps, improving critical aspects for the targeted segmentation task while ignoring unnecessary details. The selective nature of attention gates, guided by the data, facilitates a more concentrated feature representation, supporting the network's ability to model and deliver enhanced performance in semantic segmentation.

Strategic inclusion of attentive skip connections in the DAU-FI Net architecture (bottom right of Fig. \ref{fig:whole_archi}) enhances the model's capabilities by improving segmentation accuracy and reconstruction fidelity. Represented as solid ash lines, these skip connections bypass layers to provide earlier feature maps to later layers, helping preserve spatial details lost during downsampling.

The effective integration of downsampling, upsampling, attention modeling, and skip connections greatly improves both the accuracy of segmentation and the capability to reconstruct features. 

\subsection{Dual Attentive Block}
At the core of the DAU-FI Net is the Dual Attentive Block (DAB), an innovative fusion approach that integrates a spectrum of local patterns and global dependencies across multiple scales and shapes to enhance segmentation. Specifically, the DAB combines two key components: a multiscale depthwise separable convolution block and a modified concurrent spatial-channel squeeze-and-excitation (scSE) attention mechanism.

\begin{figure*}[ht]
    \centering
    \includegraphics[width=1.0\linewidth]{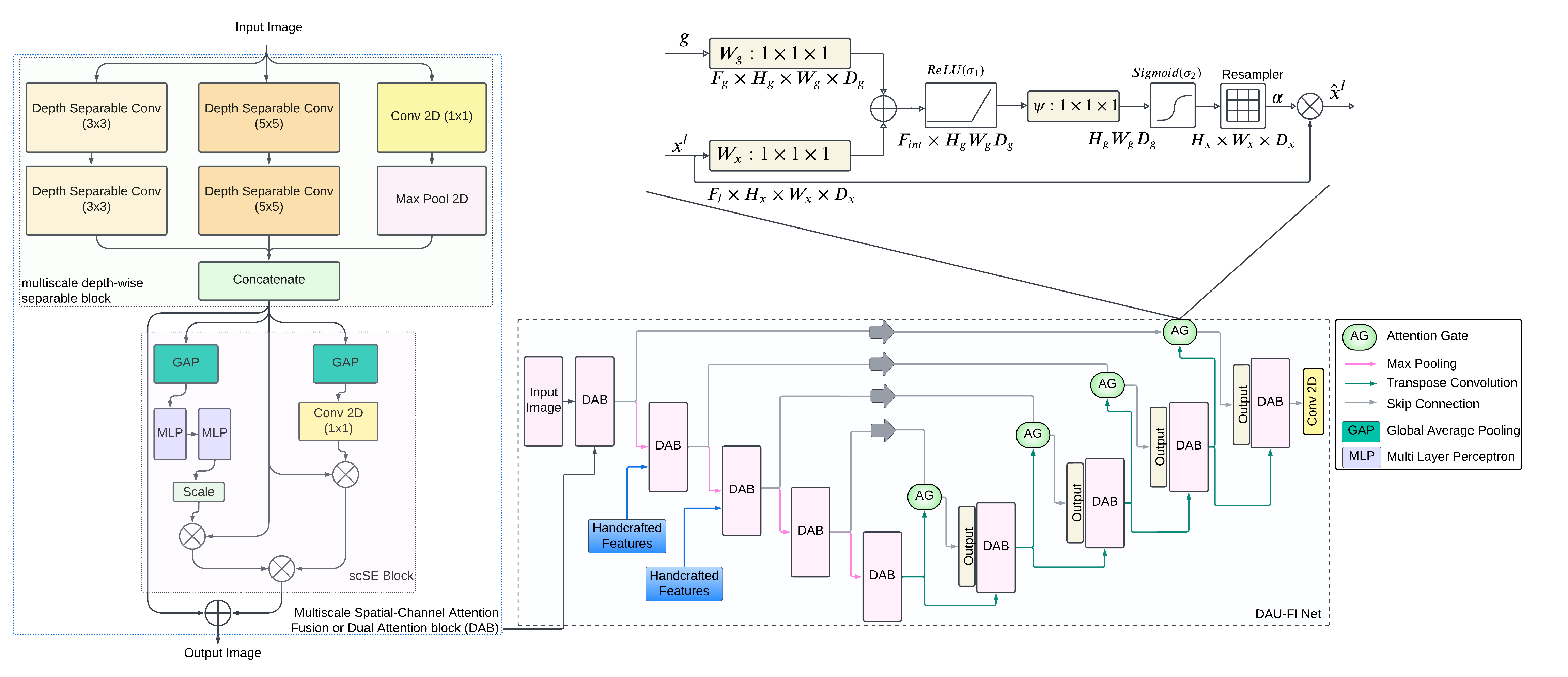}
    \caption{The Proposed Dual Attentive U-Net with Feature Infusion (DAU-FI Net) Architecture
}
\label{fig:whole_archi}
\end{figure*}

The depthwise separable convolution block, visualized in the top left of Fig. \ref{fig:whole_archi}, employs convolution kernels of varying sizes such as $3\times3$ and $5\times5$, to match the dimensions of different objects. This multiscale filtering adapts to scale variations, optimizing feature extraction across objects of diverse sizes.

Meanwhile, the concurrent scSE attention component, shown in the bottom left of Fig. \ref{fig:whole_archi}, consists of channel squeeze-and-excitation (CSE) and spatial squeeze-and-excitation (SSE) units for targeted feature recalibration. CSE emphasizes informative channels via channel-wise dimensionality reduction and global averaging. SSE generates spatial attention maps through spatial dimensionality reduction and global pooling to highlight significant regions.

Our key modification is making the reduction ratios in both CSE and SSE dynamic and learnable rather than fixed. This enables adaptive, input-specific recalibration that optimizes information flow through the block. The fusion of optimized multiscale filtering and recalibrated attention bridges the gap between fine-grained pixel intricacies and broader channel characteristics.

As visualized in Fig. \ref{fig:whole_archi}, the concurrent scSE output is merged with the initial multiscale block output by addition. This two-phase approach unifies complementary spatial-channel attention across scales into a single unified block. Overall, the DAB represents an integral innovation that improves segmentation by effectively incorporating local patterns, contextual relationships, and scaling dynamics through a cohesive fusion mechanism.

\subsection{Strategic Feature Augmentation}\label{subsec:feature-engineering}
While the DAU-FI Net architecture demonstrates promising segmentation capabilities, further refinements are required to handle multiclass imbalanced datasets with limited training samples. To address these challenges, we propose strategically augmenting the model with engineered features based on domain knowledge. Although deep neural networks demonstrate remarkable feature learning capabilities \cite{krizhevsky2012imagenet}, combining this with handcrafted input expands the feature space, providing additional guidance to improve performance when data are scarce or skewed. Our hybrid approach effectively fuses the complementary strengths of deep learning and specialized feature design.

\begin{figure}[ht]
    \centering
    \includegraphics[width=0.96\linewidth]{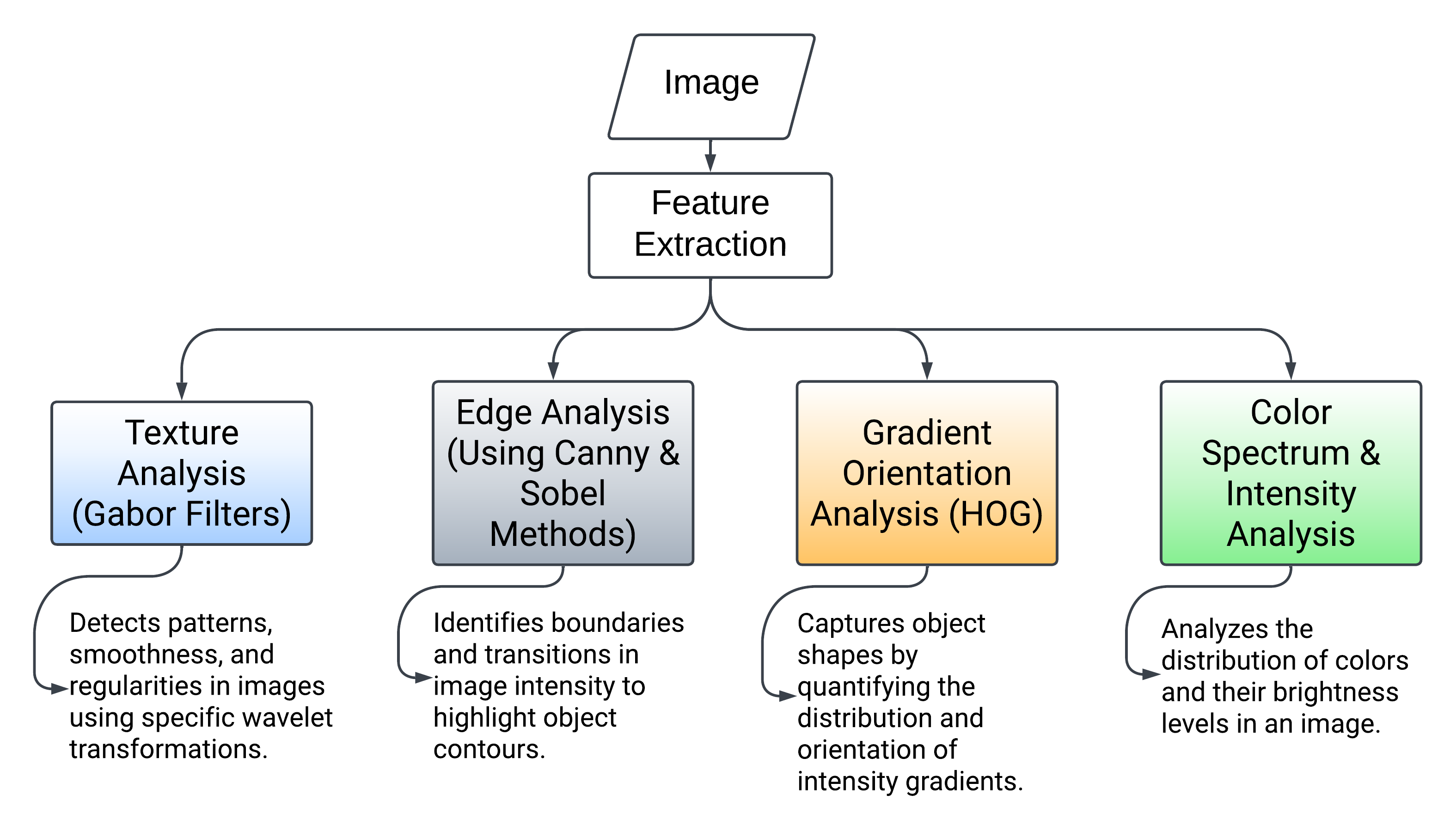}
    \caption{Feature Extraction Methods and Filters.}
    \label{fig:var_features}
\end{figure}

Specifically, we employ a comprehensive four-pronged approach utilizing: 1) Gabor filters for texture analysis to detect patterns, smoothness and irregularities via wavelet transformations; 2) Canny and Sobel edge detectors to identify boundaries and intensity transitions that highlight object contours; 3) Histogram of oriented gradients (HOG) to capture morphological attributes by quantifying the distribution and orientation of intensity gradients; and 4) Color spectrum and intensity analysis to assess the distribution and variations of colors within images. As shown in Fig. \ref{fig:var_features}, these four complementary techniques extract textural, edge-based, shape-based, and color/intensity characteristics. Fusion of these engineered inputs significantly expands the feature space beyond what deep learning alone can extract from the constrained pipe inspection data. This enables the model to overcome sample size and class imbalance limitations by using specialized handcrafted representations to boost segmentation capabilities. 

Gabor filters extract multiscale, multi-orientation textural features to robustly characterize corrosion, fractures, and blockages \cite{daugman1985uncertainty,nishikawa2012concrete}, while Sobel and Canny operators \cite{sobel2022sobel,canny1986computational} capture informative edge patterns to localize defects precisely - the adaptation of these techniques provides additional cues to distinguish complex sewer pipe deficiencies in the dataset. The \textbf{supplementary documen}t contains information about the Gabor filter-based textural analysis and the Sobel and Canny operator-based edge analysis.

\begin{figure}[ht]
    \centering
    \includegraphics[width=\columnwidth]{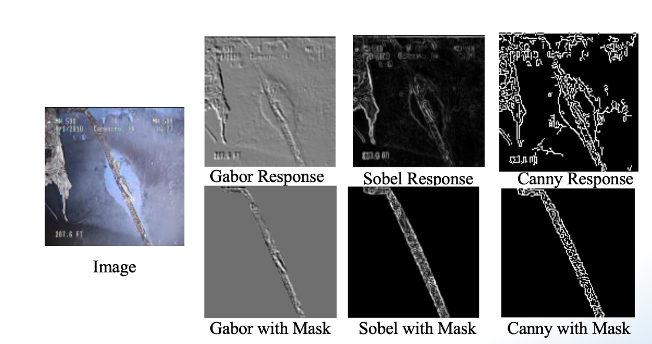}
    \caption{Image enhancement using filtering and region-based feature extraction. Top row shows the original image and results of Gabor, Sobel, and Canny filters. The bottom row presents outcomes of region-based methods, demonstrating refined representation achieved by combining filters and annotated masks}
    \label{fig:filt_image}
\end{figure}

Fig. \ref{fig:filt_image} displays the responses of the mentioned filters and the outcomes obtained by applying region-based feature extraction methods to a sample from our dataset.

\subsubsection{Gradient Orientation Analysis}
Beyond textural and edge-based features, we also explore the histogram of oriented gradient (HOG) features to characterize local shape properties and spatial arrangements of defects \cite{dalal2005histograms}. HOG analyzes gradient orientation in localized portions of images, enabling robust feature extraction useful for object detection \cite{felzenszwalb2009object}. For our sewer pipe images, HOG can help capture distinctive shape profiles of certain defects like fractures, holes, and detached joints which exhibit oriented gradient patterns. This strengthens the model's capabilities in detecting shape-based anomalies and distinguishing between different defect types exhibiting oriented gradient patterns.  

\subsubsection{Color Spectrum and Intensity Analysis}
Furthermore, color and intensity features play a vital role in image analysis. We implement color spectrum intensity analysis to examine the distribution and variations of colors within pipe images \cite{khanal2022using}. Depending on factors like corrosion and mineral deposits, defects exhibit distinctive color characteristics compared to intact regions. Capturing color and intensity features can help identify these anomalous patterns. The fusion these handcrafted features expands the representational power beyond what deep learning alone can extract from our constrained pipe inspection data.

\subsubsection{Injection of Augmented Features}
\begin{figure} [ht]
    \centering
   \includegraphics[width=0.99\linewidth]{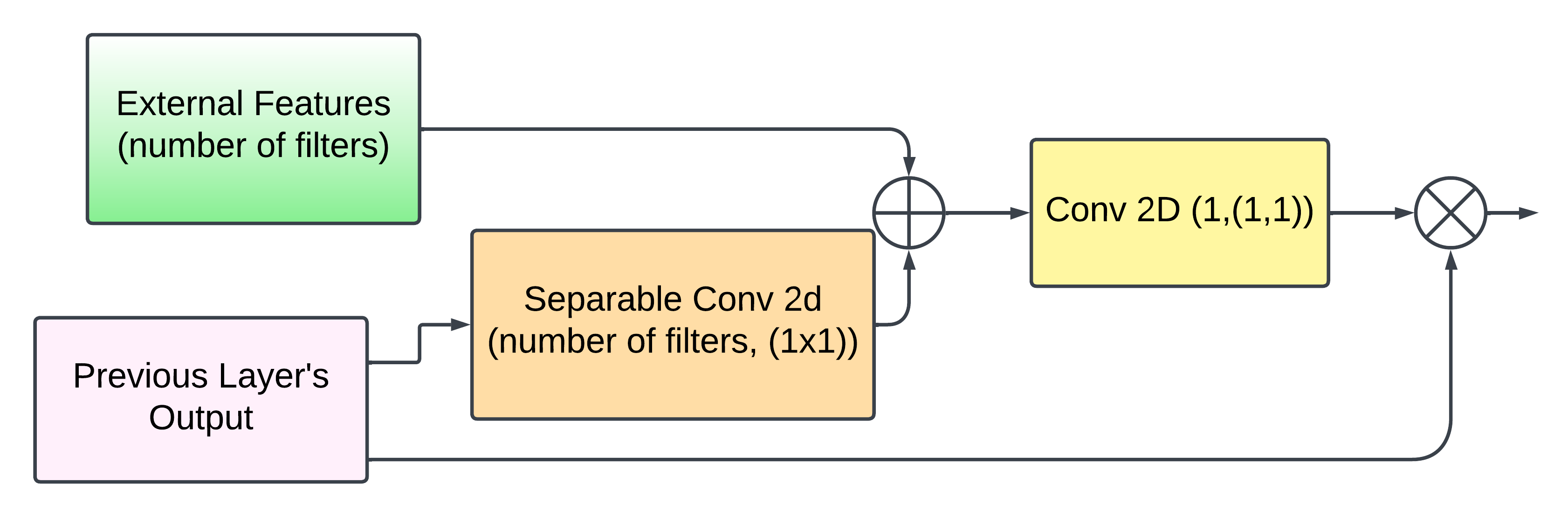}
    \caption{ Feature injection into DAU segmentation model. The process involves aligning extracted features with model layers, combining via addition, and integrating using 1x1 convolutions.}
    \label{fig:featu_inj_ways}
\end{figure}

After feature extraction, the subsequent crucial step involves integrating these features into our segmentation model, DAU. We conducted an in-depth examination of feature injection at multiple layers within the model, experimenting with various operations such as addition and multiplication. After a series of experiments, we identified the optimal approach for feature injection, which involved introducing the features into the first two layers of the model using the following method:

To begin, we aligned our extracted features, which possessed a specific number of filters, with the model's corresponding layers. This alignment necessitated the creation of convolutional layers with a matching number of channels, mirroring the number of filters or feature extractors. Subsequently, we combined these layers by addition, followed by the application of 1x1 convolutional layers with a single filter. This final step allowed us to perform element-wise multiplication with the model's input data, effectively integrating the extracted features into the segmentation process. The method is illustrated in Fig. \ref{fig:featu_inj_ways}. 

\subsubsection{Annotation-Guided Extraction}
To further optimize feature extraction, we leverage pixel-wise annotated masks from our dataset as guides for region-based filtering. We first apply selected filters to images to generate response maps. Subsequently, we perform element-wise multiplication between the response maps and annotation masks. This masks out irrelevant features while selectively preserving defect-related patterns, as determined by domain knowledge.

By carefully incorporating domain insights, feature design, and dataset annotations strengthens, we enhance the efficiency of our model in identifying and classifying pipe deficiencies. The domain guides appropriate filter selection while the annotations localize critical regions, together enabling more focused feature extraction. We empirically evaluate injecting engineered features at multiple model layers to determine optimal injection strategies. Our goal is to effectively combine the complementary strengths of deep representation learning and customized feature engineering to improve performance. The specialized feature augmentation aims to overcome data constraints, customizing the model for real-world sewer inspection deployment.

\section{Dataset Description}
This section details the methodology for creating our Sewer and Culvert Defect Segmentation Dataset. We first outline the data collection process, describing how the source videos containing labeled defect instances were obtained and pre-processed to extract key frames. Next, we discuss the pixel-wise annotation strategy used to generate fine-grained ground truth masks for enabling semantic segmentation. Technicians meticulously traced precise boundaries for each defect in the extracted frames to produce pixel-level labels. Finally, we assess the model's reliability and generalizability by evaluating performance on a public benchmark dataset for nuclei segmentation. 

\subsection{Data Collection Methodology}
We collected 580 annotated underground infrastructure inspection videos from two industry sources to assemble a robust dataset covering various real-world conditions. The videos cover both sewer pipes and culverts, introducing variations in materials, shapes, dimensions, and imaging environments. Technicians meticulously annotated each video, identifying bounding boxes and timestamps for nine common structural deficiency classes based on industry standards.

To guide model training, a professional civil engineer assigned each deficiency class an importance weight between 0 and 1, reflecting the economic and safety impacts based on US industry standards. Weights were normalized by dividing individual scores by the highest value, establishing priorities during learning.

The compiled dataset consists of a wide range of materials, shapes, and measurements found in sewer pipes and culverts. This diversity within the dataset accurately reflects the inherent variations encountered during real-life inspections of sewers and culverts in the field. It presents an added challenge due to integrating data from various sources and different structures, including culverts and sewer pipes.

\begin{table}[ht]
    \centering
    \caption{Sewer-Culvert Inspection Classes: Deficiency and Corresponding Class Importance Weights (CIW).}
    \label{tab:ninedefects}
    \begin{tabular}{lc}
        \toprule
        \textbf{Deficiency} & \textbf{CIW} \\
        \midrule
        Water Level & 0.0310 \\
        Cracks & 1.0000 \\
        Roots & 1.0000 \\
        Holes & 1.0000 \\
        Joint Problems & 0.6419 \\
        Deformation & 0.1622 \\
        Fracture & 0.5100 \\
        Encrustation/Deposits & 0.3518 \\
        Loose Gasket & 0.5419 \\
        \bottomrule
    \end{tabular}
\end{table}

\subsection{Pixel-Level Annotation}
To construct our dataset, we began by segmenting each video into individual frames, capturing one frame at predefined intervals, ranging from 4 to 10 seconds at each class annotation point within the sewer and culvert inspection video. Each annotation corresponds to a deficiency label assigned to a specific class and was timestamped to a particular second in the video. Additionally, we record the associated location within the pipe for each annotation. At present, the dataset comprises approximately 5,000 frames, including the nine deficiencies listed in Table \ref{tab:ninedefects}.

An essential aspect of our dataset preparation involved manual pixel-wise annotation to facilitate semantic segmentation. This involved accurately outlining each occurrence of a deficiency within the video frames. Highly skilled annotators carefully outlined the boundaries of each deficiency in each frame to create pixel-level masks. These masks serve as ground truth data for training and evaluating our semantic segmentation models. The pixel-wise annotation process ensures that our dataset provides the detailed information necessary to accurately identify and classify deficiencies at the pixel level, allowing the development and evaluation of robust semantic segmentation algorithms. Each of the nine annotated classes is assigned a specific color based on the structural color-coding guidelines outlined in the US NASSCO’s pipeline assessment certification program (PACP) \cite{pottawatomie}. 

\subsection{Benchmark Evaluation}


In addition to our sewer-culvert defect dataset, we also evaluate performance on the 2018 Data Science Bowl dataset for cell nuclei segmentation \cite{caicedo2019nucleus}. This dataset comprised approximately 700 segmented cell images in its initial stage, each obtained under a wide array of conditions, including different cell types, varying magnifications, and diverse imaging modalities. While this biomedical application area differs considerably from infrastructure inspection, assessing performance on an established benchmark with unique imaging characteristics still provides useful insights. The key aspects we aim to validate are the model's capability to:

\begin{itemize}
    \item Generalize to diverse datasets beyond the domain it is developed for
    \item Segment fine-grained structures from images with distinct artifacts, noise, and other complexities
    \item Handle multi-class segmentation tasks with varying numbers of output classes
\end{itemize}
Analyzing results on the cell nuclei data allows us to rigorously evaluate the robustness and limitations of our approach on a standard dataset with distinct challenges and imaging modalities. This helps identify aspects that translate across domains versus those that may require customization. The goal is to thoroughly evaluate the model to inform future development and application best.

\section{Model Training and Evaluation}
This section outlines the optimization, regularization, loss functions, and evaluation metrics used to train and rigorously evaluate our semantic segmentation models  effectively.

\subsection{Training Optimization and Regularization}
\subsubsection{Loss Function}
We use Categorical Cross-Entropy Loss during model training to enable optimized multi-class pixel classification. This commonly adopted function is well-suited for optimizing multi-class predictions in semantic segmentation tasks. It effectively minimizes the error between predicted class probabilities and ground-truth labels.

\subsubsection{Dropout Regularization}
To balance maximizing model performance and preventing overfitting, we incorporate a dropout layer before the output layer with a rate of 0.2 determined through hyperparameter tuning. This regularization technique temporarily disables random neurons during training to prevent co-adaptation. This improves generalization capabilities.

\subsubsection{Optimization Algorithm}
We utilize the Adam algorithm, which adapts the learning rates for each parameter. An initial learning rate of 0.001 and exponential decay of 10\textsuperscript{-4} after 10 epochs are used to gradually refine the model weights.

\subsection{Evaluation Metrics}
The evaluation of the model's performance on semantic segmentation of sewer and culvert defects is conducted using four key metrics: 1) Intersection over Union (IoU) with Standard and Frequency-Weighted variants to measure prediction overlap with actual data; 2) F1 Score, combining precision and recall for a balanced metric sensitive to data imbalances; 3) Balanced Accuracy, which averages recall across all classes for equitable representation; and 4) Matthew's Correlation Coefficient (MCC), gauging the quality of binary classifications, with higher scores indicating superior performance, particularly useful for datasets with skewed distributions. These evaluation metrics are elaborated in the \textbf{supplementary document}.

\section{Results}
This section presents the quantitative results from comprehensive experiments evaluating our proposed DAU-FI Net model on the challenging sewer and culvert defect dataset. We assess overall performance by benchmarking against established architectures and analyze the impact of our core innovations through detailed ablation studies.

\subsection{Comparative Evaluation}
To validate our proposed model, we performed comprehensive comparisons against prominent baseline and state-of-the-art architectures such as 1) U-Net: The pioneering fully convolutional network for segmentation; 2) Attention U-Net: U-Net with added attention modules; 3) CBAM U-Net: U-Net with concurrent spatial and channel attention blocks; 4) ASCU-Net: U-Net with a tripartite attention mechanism; and 5) Depthwise Separable U-Net (DWS MF U-Net): U-Net with multiscale filtering. We evaluated a range of metrics - IoU, FWIoU, F1-Score, Balanced Accuracy, and MCC. Table \ref{tab:Performance} summarizes the key results. 

\begin{table}[ht]
\centering
\caption{Performance Comparison of Various Models on Sewer-Culvert Defects Dataset (v1: Proposed Model with Original seSC Block; v2: Proposed Model with Improved seSC Block, w/bg: with background, w/o bg: without background)}
\label{tab:Performance}
\resizebox{\columnwidth}{!}{%
\begin{tabular}{
  @{}
  l
  S[table-format=1.5]
  S[table-format=1.5]
  S[table-format=1.5]
  S[table-format=1.5]
  S[table-format=1.5]
  S[table-format=1.5]
  @{}
}
\toprule
\textbf{Model} & {\textbf{IOU w/ bg}} & {\textbf{IoU w/o bg}} & {\textbf{FWIoU}} & {\textbf{F1}} & {\textbf{Bal. Acc}} & {\textbf{MCC}} \\
\midrule
U-Net & 0.58559 & 0.53906 & 0.48980 & 0.69333 & 0.63078 & 0.40457 \\
Attention U-Net & 0.48050 & 0.41886 & 0.43463 & 0.56858 & 0.60838 & 0.57219 \\
CBAM U-Net & 0.60501 & 0.55889 & 0.67053 & 0.71269 & 0.67964 & 0.71296 \\
ASCU-Net & 0.70358 & 0.67021 & 0.71491 & 0.81161 & 0.79463 & 0.79451 \\
DWS MF U-Net & 0.75552 & 0.72762 & 0.77249 & 0.85494 & 0.84290 & 0.84032 \\
Ours (v1) & 0.73080 & 0.70016 & 0.76035 & 0.83405 & 0.82765 & 0.81913 \\
Ours (v2) & \textbf{0.75929} & \textbf{0.73200} & \textbf{0.78054} & \textbf{0.85132} & \textbf{0.85657} & \textbf{0.84139} \\
\bottomrule
\end{tabular}
}
\end{table}

\begin{figure}[ht]
    \centering
    \includegraphics[width=\columnwidth]{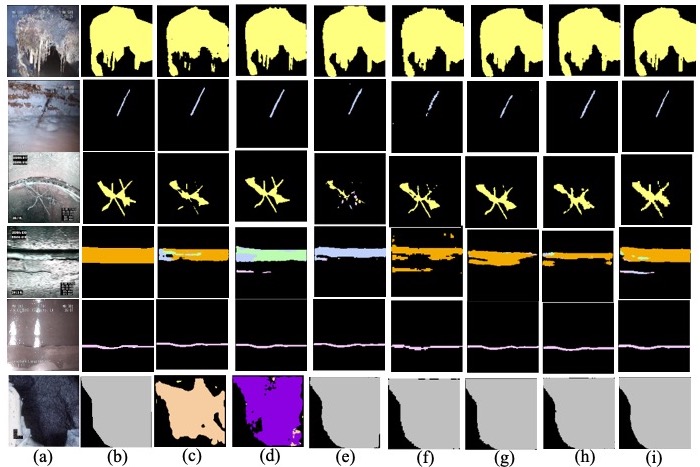}
    \caption{Visual comparisons among the segmentation models on the sewer-culvert defects dataset: (a) the original images; (b) the ground truth; (c) U-Net; (d) Attention U-Net; (e) CBAM U-Net; (f) ASCU-Net; (g) depth-wise separable U-Net with multiscale filters; (h) the proposed model with the original seSC block; and (i) DAU-FI Net with improved seSC block.}
    \label{fig:sewer_pipes_comparison}
\end{figure}

In our study, the enhanced DAU-FI Net model, particularly with the upgraded seSC Block, outperformed its advanced counterparts across all evaluated metrics. This achievement is notable given the model's reduced parameter count, as detailed in Table \ref{tab:parameters}. For a practical illustration, Fig. \ref{fig:sewer_pipes_comparison} offers a side-by-side comparison of segmentation results on sample images from our sewer-culvert defects dataset. This figure includes the original image, the ground truth mask, and predictions from various models: U-Net, Attention U-Net, CBAM U-Net, ASCU-Net, Depthwise Separable U-Net, and our DAU-FI Net. Visually, the DAU-FI Net displays more accurate defect identification and better alignment with the ground truth than its competitors.

\begin{figure*}[ht]
    \centering
    \includegraphics[width=0.78\textwidth]{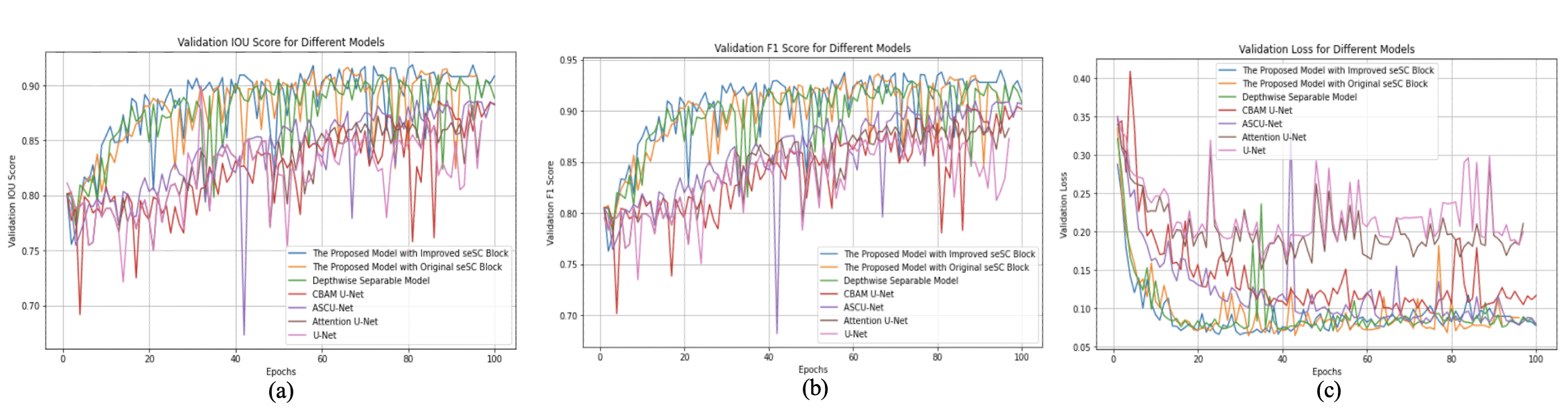}
    \caption{Validation graphs of the segmentation models: (a) the IoU validation graph; and (b) the F1-Score validation graph. (c) represents the validation loss of the sewer-culvert defects dataset. The blue line indicates the proposed model exhibits the highest validation scores compared to the other models.}
    \label{fig:validation_graphs}

\end{figure*}

Further quantitative evidence of the DAU-FI Net's efficacy is presented in Fig. \ref{fig:validation_graphs}. Here, we show the model's performance over training epochs using the sewer-culvert dataset. Specifically, Fig. \ref{fig:validation_graphs}(a) reveals the Intersection over Union (IoU) scores, with our DAU-FI Net (represented by the blue line) achieving the highest validation IoU. Similarly, Fig. \ref{fig:validation_graphs}(b) charts the F1-Score trends, again underscoring the DAU-FI Net's leading performance. Lastly, Fig. \ref{fig:validation_graphs}(c) demonstrates the validation loss trends, where our model shows faster convergence and a lower validation loss than its counterparts. These graphs collectively underscore the DAU-FI Net's enhanced segmentation accuracy.

\begin{table}[h]
\centering
\caption{Comparisons of the Number of Trainable Parameters for the Segmentation Models}
\label{tab:parameters}
\resizebox{\columnwidth}{!}{
\begin{tabular}{@{}lc@{}}
\toprule
\textbf{Model} & \textbf{Number of Parameters} \\
\midrule
Depth-Wise Separable U-Net with Multiscale Filters & 983,335 \\
DAU-FI Net (ours) & 1,456,961 \\
U-Net & 31,032,521 \\
CBAM U-Net & 31,221,065 \\
Attention U-Net & 32,430,687 \\
ASCU-Net & 31,841,202\\ 
\bottomrule
\end{tabular}}
\end{table}

\subsection{Ablation Studies}

We conducted ablation studies to analyze the impact of our core innovations - the dual attention blocks and strategic feature injection.

Integrating engineered features yielded significant IoU improvements on both the sewer-culvert and cell nuclei datasets, highlighting the efficacy of our approach. Table \ref{tab:feature} summarizes the results.

Additionally, we performed a series of experiments with different configurations (P1-P6) to determine optimal block placement, as documented in Table \ref{tab:modelsp1top5}. The best architecture was P6 which included dual attention in both encoder and decoder pathways alongside attention gates on skip connections. This configuration demonstrated the highest segmentation accuracy as seen in Table \ref{tab:ablation_study}.

\begin{table}[ht]
\centering
\caption{The effect of Feature Engineering on the Models using the sewer-culvert defects dataset}
\label{tab:feature}
\resizebox{\columnwidth}{!}{%
\begin{tabular}{
  >{\raggedright\arraybackslash}p{4cm}
  S[table-format=1.5]
  S[table-format=1.5]
  S[table-format=1.5]
  S[table-format=1.5]
  S[table-format=1.5]
  S[table-format=1.5]
}
\toprule
\textbf{Model} & {\textbf{IOU w/ bg}} & {\textbf{IoU w/o bg}} & {\textbf{FWIoU}} & {\textbf{F1-score}} & {\textbf{Bal. Acc}} & {\textbf{MCC}} \\
\midrule
Including two features on one encoder's layer (v1) & 0.89397 & 0.90525 & 0.912206 & 0.94491 & 0.94649 & {--} \\
Including two features on one encoder's layer (v2) & 0.90525 & 0.89397 & 0.912206 & 0.94491 & 0.94649 & 0.91230 \\
Including two features on two encoder's layers & 0.9239 & 0.92097 & 0.94181 & 0.95167 & 0.9613 & 0.93471 \\
U-Net & 0.78546 & 0.75889 & 0.83164 & 0.85241 & 0.84546 & 0.86430 \\
\textbf{The Proposed Model with Improved seSC Block} & 0.95679 & 0.95161 & 0.96004 & 0.97759 & 0.97478 & 0.97638 \\

\bottomrule
\end{tabular}
}
\end{table}

\begin{figure}[ht]
    \centering
    \includegraphics[width=1.0\columnwidth]{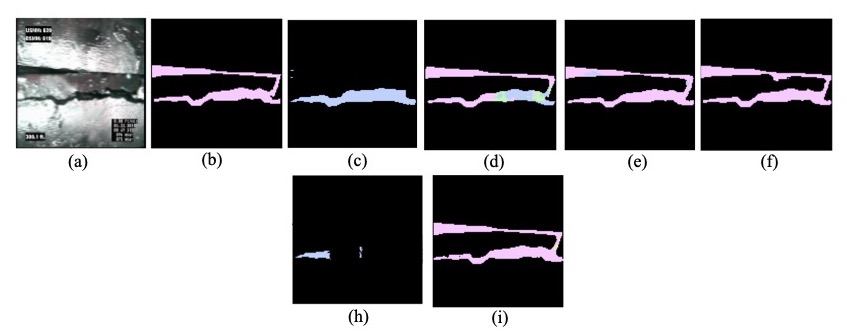}
    \caption{(a) represents the original image. (b) shows the ground truth mask. (c) displays the model's prediction without any additional features, where the model detected part of the deficiency but misclassified it, incorrectly identifying a fracture as a crack. (d) presents the model prediction after adding two features to one encoder's layer. (e) illustrates the model prediction with two features added across two encoder's layers. (f) shows the model prediction with three features added across two encoder's layers. (h) demonstrates the U-Net model prediction without any added features, and (i) reveals the U-Net prediction after incorporating three features across two encoder's layers.}
    \label{fig:enter-label}
\end{figure}

\begin{table}[ht]
\centering
\caption{The effect of feature engineering on the models using the cell nuclei dataset (v1: without feature injection; v2: with three features injected on two encoder's layers)}
\label{tab:model_comparison}
\resizebox{\columnwidth}{!}{
\begin{tabular}{>{\centering\arraybackslash}m{5cm} >{\centering\arraybackslash}m{1.5cm} >{\centering\arraybackslash}m{1.5cm} >{\centering\arraybackslash}m{1.5cm} >{\centering\arraybackslash}m{1.5cm} >{\centering\arraybackslash}m{1.5cm}}
\toprule
\textbf{Models} & \textbf{IOU w/ bg} & \textbf{IOU w/o bg} & \textbf{F1-Score} & \textbf{Balance Acc} & \textbf{MCC} \\
\midrule
Proposed model (v1) & 0.90317 & 0.83672 & 0.94783 & 0.94588 & 0.90284 \\
\textbf{Proposed model (v2)} & \textbf{0.98808} & \textbf{0.97953} & \textbf{0.99398} & \textbf{0.99423} & \textbf{0.94840} \\
\bottomrule
\end{tabular}
}
\end{table}

\begin{table}[htbp]
    \centering
    \caption{Models and Their Descriptions – A study to identify the most effective block placement within the network for performance optimization.}
    \label{tab:modelsp1top5}
    \begin{tabularx}{\columnwidth}{@{}cX@{}}
        \toprule
        Model & Description \\ 
        \midrule
        P1 & Dual Attention Block on the skip connections without including attention gates. \\
        P2 & Dual Attention Block on the decoder part without attention gates on skip connections. \\
        P3 & Dual Attention Block on the decoder part with attention gates on skip connections. \\
        P4 & Dual Attention Block on the encoder part with attention gates on skip connections. \\
        P5 & Dual Attention Block on both encoder-decoder part without attention gates on skip connections. \\
        P6 & Dual Attention Block on both encoder-decoder part with attention gates on skip connections. \\
        \bottomrule
    \end{tabularx}
\end{table}

\begin{table}[htbp]
\centering
\caption{Performance metrics of different models in the ablation study to find the optimal position of the proposed block.}
\label{tab:ablation_study}
\resizebox{\columnwidth}{!}{%
\begin{tabular}{
  l
  S[table-format=1.5]
  S[table-format=1.5]
  S[table-format=1.5]
  S[table-format=1.5]
  S[table-format=1.5]
  S[table-format=1.5]
}
\toprule
\textbf{Model} & {\textbf{IOU back}} & {\textbf{IoU No back}} & {\textbf{FWIoU}} & {\textbf{F1-score}} & {\textbf{Balanced Acc}} & {\textbf{MCC}} \\
\midrule
P1 & 0.74473 & 0.72972 & \textbf{0.78069} & 0.79106 & 0.81517 & 0.79051 \\
P2 & 0.72094 & 0.71298 & 0.69525 & 0.78098 & 0.79444 & 0.76258 \\
P3 & 0.72094 & 0.71298 & 0.74525 & 0.80217 & 0.81444 & 0.82258 \\
P4 & 0.74165 & 0.72763 & 0.76298 & 0.84944 & 0.82290 & 0.81232 \\
P5 & 0.73066 & 0.71461 & 0.76576 & 0.82770 & 0.83549 & 0.82199 \\
P6 & \textbf{0.75929} & \textbf{0.73200} & 0.78054 & \textbf{0.85132} & \textbf{0.85657} & \textbf{0.84139} \\
\bottomrule
\end{tabular}
}
\end{table}

\section{Discussion}

This section provides in-depth analysis and contextualization of our results to highlight the excellence and novel contributions of the proposed DAU-FI Net model.

The comparative benchmarking experiments validate DAU-FI Net's superior performance over sophisticated state-of-the-art architectures like Attention U-Net and CBAM U-Net. As seen in Table \ref{tab:Performance}, our model achieved the highest scores across all key metrics on the challenging sewer-culvert defect dataset.

Critically, this excellence was attained with substantially fewer parameters compared to counterparts, as noted in Table \ref{tab:parameters}. With only 1.46 million parameters, DAU-FI Net significantly reduces computations while improving accuracy. This optimization enables efficient deployment in real-world applications.

Furthermore, our ablation studies in Section VI.B confirm the value of our core innovations. Integrating engineered features provided remarkable IoU boosts on both the sewer-culvert and cell nuclei datasets as documented in Table \ref{tab:model_comparison}. This demonstrates our feature augmentation method's versatility across domains.

Additionally, the ablation experiments identified the optimal architecture configuration P6, which strategically combines dual attention blocks in the encoder-decoder pathways with attention gates on skip connections. As Table \ref{tab:ablation_study} indicates, this design achieved the best results by enhancing information flow and feature fusion.

To provide deeper perspective, Fig. \ref{fig:enter-label} visually compares segmentation outputs. DAU-FI Net generated precise and accurate masks for challenging cases, outperforming other methods. Our model detected deficiencies even when omitted from the ground truth, demonstrating robust learning.

Overall, our systematic evaluations substantiate DAU-FI Net's capabilities in handling complex multi-class segmentation tasks, even with constrained data. The framework effectively integrates complementing modalities - dual attention for representation enhancement and feature injection for expansive embeddings. This balanced approach pushes boundaries, delivering a new state-of-the-art solution.

By pioneering an optimized architecture, strategic feature augmentation, and curating a novel defect dataset, this work establishes a strong foundation for advancing real-world semantic segmentation. Our configurable methodology provides a blueprint for tackling data-scarce segmentation problems across diverse domains.

\section{Conclusion}
This study marks a significant leap in multiclass semantic segmentation, especially under the constraint of limited training data. The DAU-FI Net architecture, a cornerstone of this research, innovatively combines multiscale depth-wise separable convolutions with an advanced concurrent spatial-channel squeeze \& excitation attention unit. This integration is key to achieving nuanced segmentation by facilitating local feature learning and capturing global interdependencies.

Our rigorous testing, including a challenging sewer-culvert defects dataset and a benchmark dataset, demonstrates the robustness of DAU-FI Net. Ablation studies further reinforce the value of our approach, particularly the integration of attention mechanisms and strategic feature injections using Gabor, Sobel, and Canny filters guided by semantic masks. These techniques have yielded measurable improvements in Intersection over Union (IoU) scores across both datasets.

The methodology presented here not only enhances performance but also does so without incurring significant computational overhead. It exemplifies how domain-specific engineered features can be effectively incorporated into deep learning frameworks.

This research not only pushes the boundaries of multiclass semantic segmentation but also underscores the synergistic potential of deep learning and feature engineering, a particularly valuable insight for data-scarce scenarios. Our novel approach to adaptable feature injection and concurrent spatial-channel attention offers fresh perspectives for addressing complex segmentation tasks. The employment of the sewer-culvert defects dataset in this study adds a valuable dimension for future research explorations. The DAU-FI Net, though tested on two datasets, shows promise for various segmentation tasks. This paper lays groundwork for improved multiclass semantic segmentation architectures. Our integration of different techniques marks a significant step forward and opens doors for further research in this area.


\bibliographystyle{IEEEtran}
\bibliography{ref_dau}

\end{document}


\title{Supplementary Document for ``Dual Attention U-Net with Feature Infusion: Pushing the Boundaries of Multiclass Defect Segmentation"} 

\maketitle
This document offers a comprehensive exploration of the textural analysis using Gabor filters, as well as the edge analysis utilizing Sobel and Canny operators, which were introduced in Section III.C of the main manuscript. Additionally, it elaborates on the evaluation metrics outlined in Section V.B of the main manuscript.

\medskip
\subsection*{III.C Strategic Feature Augmentation}\label{subsec:feature-engineering}
\subsubsection{Textural Analysis}
Gabor filters are widely utilized for texture analysis and feature extraction in computer vision \cite{daugman1985uncertainty,dunn1999efficient}. Based on mathematical functions proposed by Denis Gabor in 1946, these filters capture local spatial frequency information representing texture patterns \cite{gabor1946theory,jain1991unsupervised}.

Defects in our dataset exhibit distinctive textural characteristics based on factors like corrosion, fractures, and blockages. Gabor filters can effectively extract multi-scale, multi-orientation textural features to capture these intricacies. Prior infrastructure inspection research successfully used Gabor filters to analyze concrete and asphalt defects \cite{nishikawa2012concrete, fujita2011robust}. We adapt this technique to extract textural features from sewer pipe images, providing additional cues to distinguish complex defect patterns.

The extracted Gabor features have strong potential to improve the model's capabilities in precisely localizing and classifying pipe deficiencies based on texture. By extracting multi-scale and multi-orientation textural features, we aim to provide a robust characterization of defects exhibiting corrosion, fractures, blockages and other complex damage modalities in sewer pipes. The Gabor filters enable capturing localized frequency information to distinguish nuanced textural patterns of various deficiency types.

\subsubsection{Edge Analysis}
Beyond textures, detecting edge information is also crucial for identifying pipe defects which exhibit visual irregularities. We utilize the Sobel operator, a commonly used edge detection filter in computer vision \cite{sobel2022sobel}. The Sobel applies 3x3 convolutional kernels to approximate image gradients in the horizontal and vertical directions \cite{sobel2022sobel}. For our application, defects like cracks, fractures, and detached pipe joints exhibit clear edges distinguishing the anomalous regions. By highlighting gradients, the Sobel operator can effectively capture these edge patterns characteristic of structural deformities.

We also use the Canny operator, a robust multi-stage algorithm \cite{canny1986computational}. The Canny detector applies Gaussian smoothing to reduce noise, computes gradients via Sobel filters, performs non-maximum suppression to thin edges, and uses hysteresis thresholding to track edges \cite{canny1986computational}. The Canny operator excels at detecting true edges while minimizing false positives. For sewer pipe images, this helps accurately delineate defect boundaries in noisy environments. Canny edges provide complementary clues to Sobel edges and Gabor textures for localizing anomalous regions.

\subsection*{V.B Evaluation Metrics}
We utilize a range of metrics to thoroughly evaluate model performance on the challenging sewer and culvert defect dataset.

\subsubsection{Intersection over Union Metrics}
Intersection over Union (IoU) is a key metric that quantifies the spatial overlap between the predicted and ground truth segmentation masks. It calculates the area of intersection between the prediction and the target divided by their area of union. An IoU of 1 indicates perfect overlap while 0 denotes no overlap.

We compute two IoU variants:

\begin{itemize}
  \item \textbf{Standard IoU:} We calculate this with and without the background class to isolate accuracy in precisely segmenting defects. IoU without background focuses on foreground object segmentation performance.
  \item \textbf{Frequency-Weighted IoU (FWIoU):} This version weights each class proportionally to its dataset frequency. It provides a nuanced perspective on performance across classes with varying representation. The per-class weights are defined based on insights from the civil engineer to appropriately prioritize critical structural defects.
\end{itemize}

Together, these IoU metrics enable a comprehensive evaluation. Standard IoU assesses overall segmentation quality and accuracy. Frequency-weighted IoU accounts for dataset imbalances and domain knowledge to provide additional insight into model proficiency in identifying rare but critical defects.

\subsubsection{F1 Score}

The F1 score is defined as the harmonic mean of precision and recall, calculated as:

\begin{equation}
    F1 = 2 \times \frac{\text{precision} \times \text{recall}}{\text{precision} + \text{recall}}
\end{equation}

It provides a balanced evaluation by combining precision, which measures the accuracy of positive predictions, and recall, which quantifies the ability to detect all positive cases. Unlike metrics biased toward predicting either positive or negative instances, the F1 score accounts for both false positives and false negatives. This provides comprehensive insight when evaluating models on skewed, imbalanced defect data.

Maximizing the F1 score requires generating accurate positive predictions while also maintaining high recall across rare classes. This prevents inflated performance estimates due to dataset imbalances. The balance between precision and recall makes the F1 score a valuable metric for the assessment of segmentation model proficiency in our safety-critical application.

\subsubsection{Balanced Accuracy} 
Balanced accuracy individually computes the proportion of correct positive predictions for each class based on its recall. Then the mean of these per-class recall values is derived, essentially balancing the contribution of each class to the overall metric. By considering recall on a per-class basis before averaging, balanced accuracy prevents large classes from disproportionately dominating the measurement. This provides a more representative evaluation of model performance on our imbalanced real-world data. Conventional accuracy tends to be inflated for the majority class and underestimates performance in rare classes. In contrast, balanced accuracy gives equal weight to all categories regardless of frequency. This metric allows a comprehensive assessment essential for our defect segmentation task, where certain structural deficiencies appear much less frequently.

\subsubsection{Matthew's Correlation Coefficient (MCC)}
MCC is considered a robust metric for model evaluation that measures the correlation between predicted labels and true labels, particularly with skewed dataset distributions. By factoring in true and false positives and negatives, it provides a nuanced and comprehensive measure beyond accuracy. An MCC of 0 generally signifies a model unable to discern signal from noise. In our experiments, higher MCC values demonstrate improved model capability in distinguishing defects from intact regions across imbalanced classes. Comparison of MCCs helps to make reliable benchmarking of different segmentation architectures. Thus, along with metrics like IoU and F1 score, MCC provides a holistic perspective for rigorous model evaluation.

\bibliographystyle{IEEEtran}
\bibliography{ref_dau}